\documentclass[letterpaper, 10 pt, conference]{ieeeconf}  

\IEEEoverridecommandlockouts                              

\usepackage{amsmath,amsfonts,bm}

\def\eqref#1{equation~\ref{#1}}

\def\1{\bm{1}}

\DeclareMathAlphabet{\mathsfit}{\encodingdefault}{\sfdefault}{m}{sl}
\SetMathAlphabet{\mathsfit}{bold}{\encodingdefault}{\sfdefault}{bx}{n}

\usepackage{graphics} 
\usepackage{epsfig} 
\usepackage{mathptmx} 
\usepackage{times} 
\usepackage{amsmath} 
\usepackage{amssymb}  

\usepackage{mathtools}

\usepackage{multirow, booktabs}

\usepackage{xcolor}

\usepackage[utf8]{inputenc} 
\usepackage{pifont}
\newcommand{\cmark}{\ding{51}}%
\newcommand{\xmark}{\ding{55}}%

\usepackage{subfigure}
\usepackage{caption}
\usepackage{tabularx}
\usepackage{float}
\usepackage{epsfig}
\usepackage{epstopdf}

\usepackage{dirtytalk}

\usepackage{algorithm}
\usepackage[noend]{algpseudocode}
\usepackage{multirow}
\usepackage[section]{placeins}
\usepackage{enumerate}

\newcommand{\pname}{VERSE}

\title{\LARGE \bf
{\pname}: Virtual-Gradient Aware Streaming Lifelong Learning  \\ with Anytime Inference
}

\author{Soumya Banerjee$^{1}$, Vinay K. Verma$^{2}$, Avideep Mukherjee$^{1}$, Deepak Gupta$^{2}$, Vinay P. Namboodiri$^{3}$, Piyush Rai$^{1}$
\thanks{$^{1}$ IIT Kanpur, India,
        {\tt\small \{soumyab, avideep, piyush\}@cse.iitk.ac.in}}
\thanks{$^{2}$ Amazon, India,
        {\tt\small \{vinayugc, deepakgupta.cbs\}@gmail\newline.com}}
\thanks{$^{3}$ University Of Bath, UK,
        {\tt\small vpn22@bath.ac.uk}}
}

\begin{document}

\maketitle
\thispagestyle{empty}
\pagestyle{empty}

\begin{abstract}
  Lifelong learning or continual learning is the problem of training an AI agent continuously while also preventing it from forgetting its previously acquired knowledge. 
  Streaming lifelong learning is a challenging setting of lifelong learning with the goal of continuous learning in a dynamic non-stationary environment without forgetting. We introduce a novel approach to lifelong learning, which is streaming (observes each training example only once), requires a single pass over the data, can learn in a class-incremental manner, and can be evaluated on-the-fly (anytime inference). To accomplish these, we propose a novel \emph{virtual gradients} based approach for continual representation learning which adapts to each new example while also generalizing well on past data to prevent catastrophic forgetting. Our approach also leverages an exponential-moving-average-based semantic memory to further enhance performance. Experiments on diverse datasets with temporally correlated observations demonstrate our method's efficacy and superior performance over existing methods.
\end{abstract}


\section{INTRODUCTION}\label{sec:introduction}


Continuous machine perception is crucial for AI agents to learn while interacting with the environment, preventing catastrophic forgetting~\cite{thrun1998lifelong}. Lifelong Learning (LL) or Continual Learning (CL)~\cite{kirkpatrick2017overcoming,li2017learning} methods are designed with the goal to accomplish this. Recent CL research focuses mainly on static environments~\cite{aljundi2018memory,wu2019large,wu2018memory,kemker2017fearnet,kirkpatrick2017overcoming}, assuming large batch data, ignoring changing data distribution, and requires multiple passes over the data to facilitate CL. However, these approaches are not suitable for rapidly changing dynamic environments. While there have been efforts to enable CL in online settings~\cite{prabhu2020gdumb}, these methods have various limitations, such as batch data requirements, inability to perform anytime inference (i.e., asking the model to make predictions while it is still training), and the need for large replay buffers, limiting their applicability in Streaming Lifelong Learning (SLL)~\cite{hayes2019memory,hayes2019remind,banerjee2023streaming}. In SLL, the goal is to learn by observing each training example only once without forgetting.

\begin{table*}[ht]


  \scriptsize
  \centering
  \caption{Baseline approaches are categorized based on simplifying assumptions. $\zeta(n)$ denotes the number of times the network visits the data for continual learning. It follows that $\zeta(n) \gg \zeta(2) > \zeta(1)$ with '-' indicating we are unable to find the exact value. Categories: B (Batch), O (Online), S (Streaming).}
  \label{tab:Table_baseline_categorization}
  
  \vspace{-2mm}
   \setlength{\tabcolsep}{-0.04mm}
  
    \scriptsize

    
    \begin{tabular}{  c | c|  c|  c|  c|  c|  c|  c|  c|  c|  c|  c|  c|  c|  c|  c|  c|  c } 
      \toprule

       \textbf{Methods} & \shortstack{{LWF} \\ \cite{li2017learning}} &  \shortstack{{\;EWC++\;} \\ \cite{kirkpatrick2017overcoming,chaudhry2018riemannian}} & \shortstack{{MAS} \\ \cite{aljundi2018memory}} & \shortstack{{SI} \\ \cite{zenke2017continual}} & \shortstack{{VCL} \\ \cite{nguyen2017variational}} & \shortstack{{\;CVCL\;} \\ \cite{nguyen2017variational}} & \shortstack{{GEM} \\ \cite{lopez2017gradient}} & \shortstack{{AGEM} \\ \cite{chaudhry2018efficient}} & \shortstack{{\;GDumb\;} \\ \cite{prabhu2020gdumb}} & \shortstack{{\;TinyER\;} \\ \cite{chaudhry2019tiny}} & \shortstack{{DER} \\ \cite{buzzega2020dark}} & \shortstack{{\;DER++\;} \\ \cite{buzzega2020dark}} & \shortstack{{\;ExStream\;} \\ \cite{hayes2019memory}} & \shortstack{{\;REMIND\;} \\ \cite{hayes2019remind}} & \shortstack{{\;CLS-ER\;} \\ \cite{arani2022learning}} & \shortstack{\textbf{\;\;{\pname}\;\;} \\ \textbf{\;\;(Ours)\;\;}} & \shortstack{{\textbf{CISLL}} \\ {\;\textbf{Constraints}\;}} \\

      \cmidrule{1-18}

       \textbf{Type} & B & B & B & B & B & B & O & O & O & O & O & O & S & S & B & S &  \\

      \cmidrule{1-18}

       \shortstack{{\textbf{Batch-Size}}  {$\mathbf{(N_{t})}$\;}} & $\;N_{t} \gg 1\;$ & $\;N_{t} \gg 1\;$ & $\;N_{t} \gg 1\;$ & $\;N_{t} \gg 1\;$ & $\;N_{t} \gg 1\;$ & $\;N_{t} \gg 1\;$ & $\;N_{t} \gg 1\;$ & $\;N_{t} \gg 1\;$ & $\;N_{t} \gg 1\;$ & $\;N_{t} \gg 1\;$ & $\;N_{t} \gg 1\;$ & $\;N_{t} \gg 1\;$ & $\;N_{t} = 1\;$ & $\;N_{t} = 1\;$ & $\;N_{t} \gg 1\;$ & $\;N_{t} = 1\;$ &  \\

      \cmidrule{1-18}

      \shortstack{{\textbf{Fine-tuning}}} & \xmark & \xmark & \xmark & \xmark & \xmark & \cmark & \xmark & \xmark & \cmark & \xmark & \xmark & \xmark & \xmark & \xmark & \xmark & \xmark & \xmark \\

      \cmidrule{1-18}

      \shortstack{{\textbf{Single Pass}}} & \xmark & \xmark & \xmark & \xmark & \xmark & \xmark & \cmark & \cmark & \xmark & \cmark & \cmark & \cmark & \cmark & \cmark & \xmark & \cmark & \cmark \\

      \cmidrule{1-18}

       \shortstack{{\textbf{Follows CIL}}} & \xmark & \xmark & \xmark & \xmark & \xmark & \xmark & \xmark & \xmark & \cmark & \cmark & \cmark & \cmark & \cmark & \cmark & \cmark & \cmark & \cmark \\

      \cmidrule{1-18}

       \shortstack{{\textbf{Subset Replay}}} & n/a & n/a & n/a & n/a & n/a & \xmark & \cmark & \cmark & \xmark & \cmark & \cmark & \cmark & \xmark & \cmark & \cmark & \cmark & \cmark \\

      \cmidrule{1-18}

       \shortstack{{\textbf{Training Time\;}}} & $\zeta(n)$ & $\zeta(n)$ & $\zeta(n)$ & $\zeta(n)$ & $\zeta(n)$ & $\zeta(n)$ & $\zeta(1)$ & $\zeta(1)$ & $\zeta(1)$ & $\zeta(1)$ & $\zeta(1)$ & $\zeta(1)$ & $\zeta(1)$ & $\zeta(1)$ & $\zeta(n)$ & $\zeta(1)$ & $\zeta(1)$ \\

      \cmidrule{1-18}

       \shortstack{{\textbf{Inference Time\;}}} & $\zeta(1)$ & $\zeta(1)$ & $\zeta(1)$ & $\zeta(1)$ & $\zeta(1)$ & $\zeta(n)$ & $\zeta(1)$ & $\zeta(1)$ & $\zeta(n)$ & $\zeta(1)$ & $\zeta(1)$ & $\zeta(1)$ & $\zeta(1)$ & $\zeta(1)$ & $\zeta(1)$ & $\zeta(1)$ & $\zeta(1)$ \\

      \cmidrule{1-18}

       \shortstack{{\textbf{Buffer Size}}} & n/a & n/a & n/a & n/a & n/a & - & - & - & - & $\leq 5\%$ & $\leq 5\%$ & $\leq 5\%$ & $\leq 5\%$ & $\gg 10\%$ & $\leq 5\%$ & $\leq 5\%$ &  \\

      \cmidrule{1-18}

       \shortstack{{\textbf{Doesn't violate\;}} \\ {\;\textbf{CISLL}\;}} & {\xmark} & {\xmark} & {\xmark} & {\xmark} & {\xmark} & {\xmark} & {\xmark} & {\xmark} & {\xmark} & \cmark & \cmark & \cmark & {\xmark} & \cmark & {\xmark} & \cmark &  \\
       
      \bottomrule
    \end{tabular}
  
      \vspace{-2.4em}
      
\end{table*}

Below, we outline the key properties of SLL~\cite{hayes2019memory,banerjee2023streaming}:

\begin{itemize}

  \itemsep0em
  
  \item The AI agent observes each training example only once without storing it in memory. 
  
  \item The agent is required to adapt to new sample(s) in a single pass. 
  
  \item The input data stream may exhibit temporal correlations, deviating from the typical i.i.d pattern. 
  
  \item The agent is required to be evaluated at any time (anytime inference) \footnote{Note that this is different from the \emph{anytime inference} considered in~\cite{ruiz2021anytime,huang2017multi,li2019improved,yang2020resolution,zhang2018graph} where the model has multiple decision/exit points, one of which is chosen at inference time depending on the desired inference latency; our setting is similar to that considered in~\cite{koh2021online,hayes2019remind}} without fine-tuning its parameters. 
  
  \item The agent needs to perform class-incremental streaming lifelong learning (CISLL), i.e., predict a class label from all the previously observed classes. 
  
  \item To make it practical, especially in resource-constrained environments, the agent should minimize its memory requirements. 

\end{itemize}


Existing CL approaches often make strong assumptions that violate one or more key constraints required for SLL. Despite being desirable because of also being closer to biological learning~\cite{hayes2019remind}, SLL hasn't received much attention. The SLL setting is natural in real-world scenarios like home robots, smart appliances, and drones, where AI agents must adapt quickly and continuously without forgetting. Table~\ref{tab:Table_baseline_categorization} categorizes existing CL approaches based on their underlying assumptions, revealing that only a few non-SLL methods can be adapted to the SLL setting without violating the constraints. Notably, ExStream~\cite{hayes2019memory}, being an SLL method, violates subset replay in SLL by using all past samples for CL. Non-SLL methods like TinyER~\cite{chaudhry2019tiny} and DER/DER++~\cite{buzzega2020dark} perform poorly when applied in the SLL setting.

\begin{figure}[t]
  \centering
  \includegraphics[width=8.2cm, height=1.8cm]{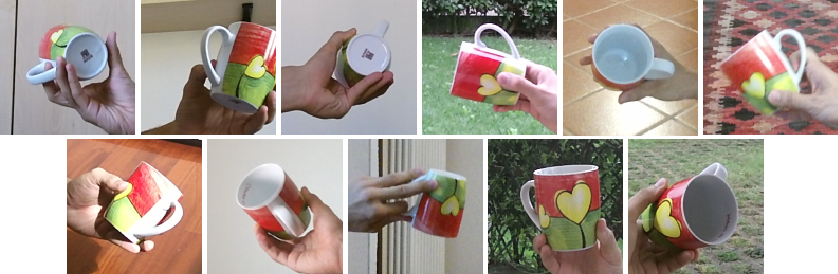}  
  \vspace{-0.2em}

  \caption{SLL involves continuous learning from non-i.i.d. labeled streams with multiple views without forgetting. This fig. shows temporally ordered cup frames from CoRe50~\cite{lomonaco2017core50}.}
  \label{fig:core50_images}
  \vspace{-2em}
\end{figure}

We introduce \textbf{V}irtual Gradi\textbf{E}nt Awa\textbf{R}e \textbf{S}treaming L\textbf{E}arning ({\pname}), a rehearsal-based CL model, facilitating CISLL in deep neural networks (DNNs). {\pname} \emph{implicitly} regularizes the network parameters by employing virtual gradient updates, fostering robust representations that require minimal changes to adapt to the new task/sample(s), preventing catastrophic forgetting. We also utilize a small episodic buffer to store past samples, which are used for both local/virtual and global parameter updates to perform a single-step virtual gradient regularization, enabling CL. {\pname} first adapts to a new example with a \emph{virtual} (local) parameter update and generalizes to past samples with a \emph{global} parameter update, promoting convergence between the two. This process facilitates SLL allowing the AI agent to be evaluated on-the-fly (anytime inference) without parameter fine-tuning with the stored samples.



Moreover, {\pname} utilizes an exponential-moving-average~\cite{tarvainen2017mean,cai2019exploring,deng2021unbiased,chen2020multi} based semantic memory akin to long-term memory in mammalian brains~\cite{thompson1994organization,linden1994long,holtmaat2009experience}. Semantic memory is updated intermittently, consolidating new knowledge within the agent's parameters. It interacts with episodic memory, interleaving the past predictions on stored buffer samples, minimizing self-distillation loss~\cite{hinton2015distilling,zhang2019your} to prevent forgetting and enhance the agent's performance.


Experimental results on three temporally contiguous datasets show that {\pname} is effective in challenging SLL scenarios. It outperforms recent SOTA methods, with ablations confirming the significance of its components.


\textbf{In summary, our contributions are as follows:}
\begin{itemize}

  \itemsep0em

  \item We present a novel approach {\pname}, a rehearsal-based virtual gradient regularization framework, that incorporates both virtual and global parameter updates to mitigate catastrophic forgetting in CISLL, enabling 'any-time-inference' without fine-tuning.

  \item We propose a semantic memory based on an exponential-moving-average approach, which enhances the agent's overall performance.

  \item Through empirical evaluations and ablation studies conducted on three benchmark datasets with temporal correlations, we affirm the superiority of {\pname} over the existing SOTA methods.

\end{itemize}
\vspace{-0.4em}

\section{RELATED WORK}\label{sec:related_work}
\vspace{-0.3em}

This section briefly summarizes different CL paradigms.


\textbf{Task Incremental Learning (TIL).} In TIL, the AI agent learns from task-batches, observing samples related to specific tasks~\cite{aljundi2017expert,aljundi2018memory,chaudhry2018riemannian,li2017learning,kirkpatrick2017overcoming,shin2017continual,zenke2017continual,nguyen2017variational,aljundi2018selfless}, each involving learning a few distinct classes. These methods rely on knowing the task-identifier during inference; otherwise, it leads to severe catastrophic forgetting~\cite{chaudhry2018riemannian}.


\textbf{Incremental Class Batch Learning (IBL).} IBL, also referred to as class incremental learning (CIL), assumes that the dataset is divided into batches, each containing samples from different classes~\cite{chaudhry2018riemannian,chaudhry2019tiny,rios2018closed,wu2019large,belouadah2019il2m,buzzega2020dark,arani2022learning,hou2019learning,tao2020topology,yoon2017lifelong,verma2021efficient}. The AI agent observes and can loop over these batches in each incremental session. During inference, the agent isn't provided with task labels and is evaluated over all the observed classes.

\textbf{Online Continual Learning (OCL).} Unlike TIL and IBL, OCL involves an AI agent sequentially observing and adapting to samples in a \emph{single pass} over the entire dataset, avoiding catastrophic forgetting~\cite{prabhu2020gdumb,chaudhry2019tiny,aljundi2019gradient,aljundi2019online,chaudhry2018efficient,lopez2017gradient}. While these methods enable continuous learning in dynamic environments, they have various limitations: $(i)$ they require data in batches, assuming $\forall t, |B_{t}| \gg 1$, where $B_{t}$ is a batch of samples at time $t$, $(ii)$ they need fine-tuning before inference, lacking any-time-inference ability (e.g., GDumb~\cite{prabhu2020gdumb}, which is an OCL method, requires fine-tuning model parameters with replay-buffer samples before each inference), $(iii)$ they require large replay buffers~\cite{hayes2019remind}.

\textbf{Streaming Lifelong Learning (SLL).} SLL, a challenging variant of LL, enables CL in a rapidly changing environment without forgetting~\cite{banerjee2023streaming,hayes2019memory,hayes2019remind}. It shares similarities with OCL but has additional constraints: $(i)$ SLL limits the batch size to one datum per incremental step, while OCL requires $|B_{t}| \gg 1$, $(ii)$ it doesn't allow AI agent to fine-tune its parameters during training or inference. Additionally, in SLL, the input data stream can be temporally correlated in terms of class instance and instance ordering. Detailed essential and desirable properties of SLL are discussed in Section~\ref{sec:introduction}.


To our knowledge, ExStream~\cite{hayes2019memory}, REMIND~\cite{hayes2019remind}, and BaSiL~\cite{banerjee2023streaming} are the three methods tackling the challenging SLL setting. However, it's important to note some of the key differences: ExStream~\cite{hayes2019memory} uses full-buffer replay, violating the subset-replay constraint; REMIND~\cite{hayes2019remind} stores a much larger number of past samples compared to other baselines (e.g., iCaRL~\cite{rebuffi2017icarl} stores 10K past ImageNet~\cite{deng2009imagenet,russakovsky2015imagenet} samples, whereas REMIND stores 1M samples); BaSiL~\cite{banerjee2023streaming} focuses on Bayesian methods for SLL and relies entirely on pretrained weights for visual features in SLL. It does not adapt convolutional layers to sequential data and only trains linear layers, potentially posing severe challenges with non-i.i.d. data. In contrast, our approach ({\pname}) adheres to the SLL constraints, stores a limited number of past samples, replays only a subset of buffer samples, and trains both convolutional and fully connected (FC) layers for SLL.


In this paper, we introduce {\pname}, adhering to CISLL constraints, enabling CL in challenging SLL settings. We compare {\pname} with SLL frameworks, REMIND~\cite{hayes2019remind} and ExStream~\cite{hayes2019memory}, as well as various OCL and IBL methods.

\vspace{-0.3em}
\section{Proposed Approach}\label{sec:proposed_method}
\vspace{-0.2em}

\begin{figure}[t]
  \centering
  \includegraphics[width=8cm, height=3.5cm]{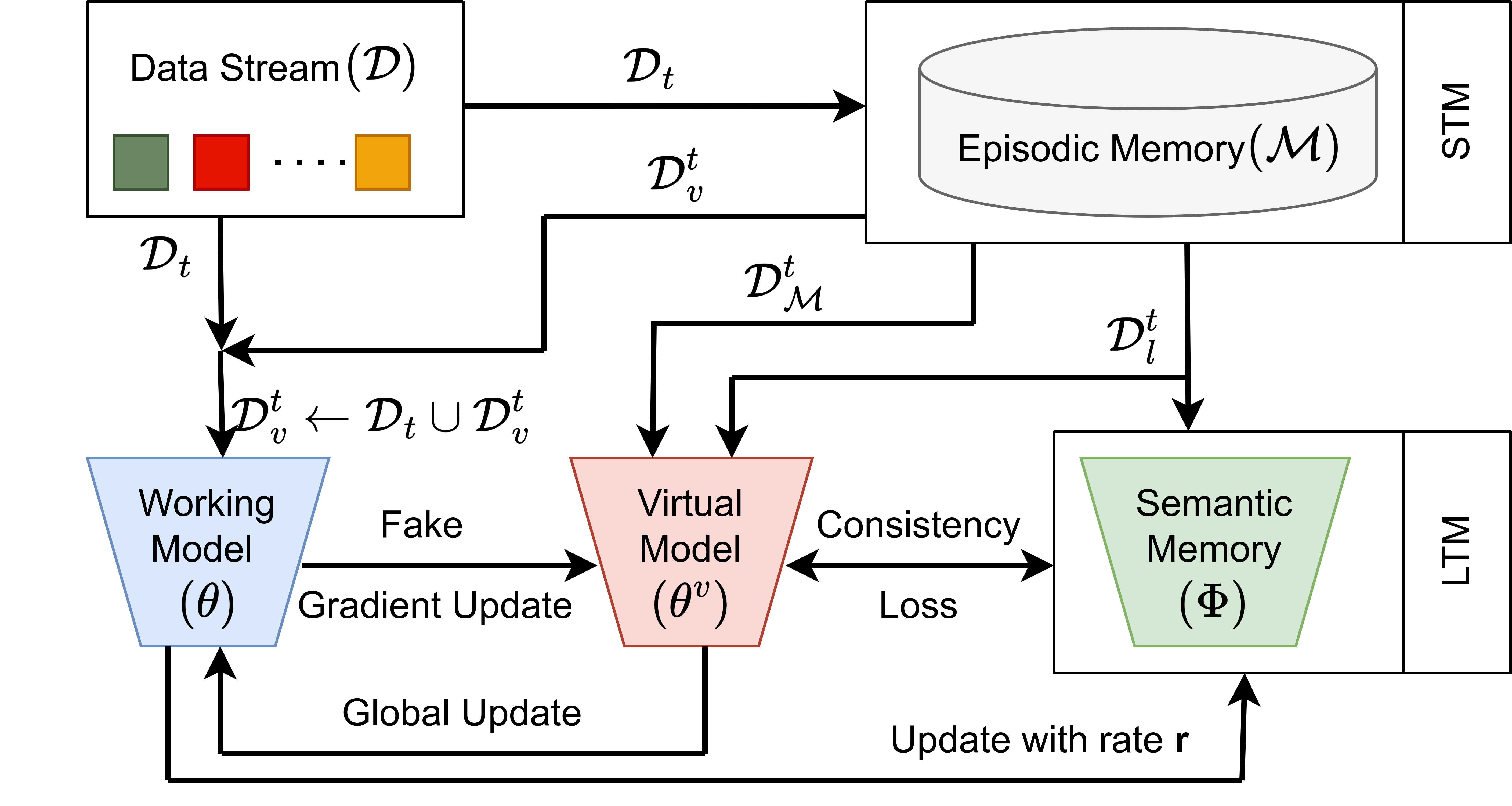}  
  \vspace{-0.2em}

  \caption{In {\pname}, Virtual-gradient-regularization (VGR) enables CL by adapting to new sample(s) with a virtual model $(\theta^{v})$, which computes the final model $(\theta)$ through rehearsal. Episodic memory (TEM) stores a few observed samples, while Semantic memory (SEM) enforces consistency with self-distillation loss, improving overall performance.}
  \label{fig:proposed_model}
  \vspace{-1.8em}
\end{figure}

In this section, we introduce the proposed approach {\pname} ({Fig.~\ref{fig:proposed_model}}), which trains the CNN architecture in CISLL setup. The model, consisting of the parameters $\Theta=\{\xi,\theta\}$, comprises of two components: $(i)$ a non-plastic feature extractor $(G_\xi)$ with parameter ${\xi}$, and $(ii)$ a plastic neural network $(F_\theta)$ with parameter $\theta$. $G_\xi$ includes the initial CNN layers, and $F_\theta$ encompasses the final few layers, including the fully connected (FC) layers. The class label output for a given input $\boldsymbol{x}$, is predicted as: $y = F_\theta(G_\xi(\boldsymbol{x}))$.


We focus on adapting the plastic network $F_\theta(\cdot)$ while keeping non-plastic parameters $\xi$ frozen throughout. In each streaming incremental step of CISLL, data arrives sequentially, $\mathcal{D}_{t} = (\boldsymbol{x}_{t}, y_{t})$, one datum at a time, and the learner adapts without catastrophic forgetting~\cite{goodfellow2013empirical,mccloskey1989catastrophic} by observing this datum only once. The following section provides brief details of the proposed model.

\subsection{Virtual Gradient as Regularizer}\label{sec:meta_regularization}
\vspace{-0.2em}

We denote by $\mathcal{D}_{1}, \mathcal{D}_{2}, \dots, \mathcal{D}_{T}$ the set of tasks, with ${(\boldsymbol{z}_{t}, y_{t})} = {(G_\xi(\boldsymbol{x}_{t}), y_{t})} \sim \mathcal{D}_{t}$ with $|\mathcal{D}{t}| = 1$, arriving one-by-one in each incremental step. During the training of the $t^{th}$ task only the data $\mathcal{D}_{t}$ is available, the previous tasks' data $\mathcal{D}_{1}, \mathcal{D}_{2}\dots \mathcal{D}_{t-1}$ are discarded, and we only keep a few samples into a small memory buffer $\mathcal{M}$. Optimizing network parameters with a single example is challenging; therefore, we select a subset of samples with size $C$ from memory $\mathcal{M}$, that is, $\mathcal{D}_{v}^{t} \sim \mathcal{M}$ with {$|\mathcal{D}_{v}^{t}| = C$}. We combine the subset $\mathcal{D}_{v}^{t}$ with the new example $\mathcal{D}_{t}$ to form a joint batch: $\mathcal{D}_{v}^{t} \leftarrow \mathcal{D}_{t} \cup \mathcal{D}_{v}^{t}$ and compute the cross-entropy loss, as defined below:
\vspace{-0.1em}
\begin{equation}\label{eq:lossnew}
    \mathcal{L}^{t}_{v} = \mathbb{E}_{(\boldsymbol{z}, y) \sim \mathcal{D}^{t}_{v}} \left[ \mathcal{L}_{CE}(y, F_{\theta}(\boldsymbol{z})) \right]
\end{equation}
Let $\nabla_{\theta}\mathcal{L}^{t}_{v}(F_{\theta})$ be the gradient of the loss (Eq.~\ref{eq:lossnew}) w.r.t. the plastic network parameters $\theta$. With this gradient, we compute the updated \emph{local parameter} as follows:
\vspace{-0.1em}
\begin{equation}\label{eq:virtualgrad}
    \theta^{v} \leftarrow \theta - \alpha \nabla_{\theta}\mathcal{L}^{t}_{v}(F_{\theta})
\end{equation}

\begin{table*}[ht]

   
   \scriptsize
   \centering


 \caption{$\boldsymbol{\Omega}_{\text{all}}$ results. Best-performing CISLL method is highlighted in \textbf{Bold}. The reported results are an average over 10 runs with different permutations of data. Offline model is trained once. $\widehat{\text{Offline}} = \frac{1}{T} \sum_{t = 1}^{T} \boldsymbol{\alpha}_{\text{Offline}, t}$, with '-' denoting experiments we are unable to run due to compatibility issues.}
 \label{Table_omega_all_results}
 \vspace{-0.5em}  
   
 \scriptsize
 
 \resizebox{\textwidth}{!}{
   \begin{tabular}{ c | c  c  c | c  c  c | c  c  c | c  c  c } 
     \toprule

     \multirow{2}{*} { \textbf{Method} } & \multicolumn{3}{c}{\textbf{iid}} & \multicolumn{3}{c}{\textbf{Class-iid}} & \multicolumn{3}{c}{\textbf{instance}} & \multicolumn{3}{c}{\textbf{Class-instance}} \\
     
     \cmidrule{2-13}
     & \shortstack{\textbf{iCub1.0}} & \shortstack{\textbf{iCub28}} & \shortstack{\textbf{CoRe50}} & \shortstack{\textbf{iCub1.0}} & \shortstack{\textbf{iCub28}} & \shortstack{\textbf{CoRe50}} & \shortstack{\textbf{iCub1.0}} & \shortstack{\textbf{iCub28}} & \shortstack{\textbf{CoRe50}} & \shortstack{\textbf{iCub1.0}} & \shortstack{\textbf{iCub28}} & \shortstack{\textbf{CoRe50}} \\
     \midrule 
     
     Fine-tune  & 0.9550    & 0.8432  & 0.9681  & 0.3902  & 0.4265  & 0.4360 & 0.1981  & 0.2483  & 0.2468  & 0.3508  & 0.4810  & 0.3400 \\
 
     EWC++~\cite{kirkpatrick2017overcoming,chaudhry2018riemannian}  &  -  & - & - & 0.3747 & 0.4218  & 0.4307 &  -  & - & - & 0.3507  & 0.4805  & 0.3401 \\
 
     MAS~\cite{aljundi2018memory}  &  -  & - & - & 0.3758  & 0.4334  & 0.4333 &  -  & - & - & 0.3509  & 0.4807  & 0.3401 \\
 
     AGEM~\cite{chaudhry2018efficient}  &  -  & - & - & 0.4626  & 0.7507  & 0.5633 & -    & - & - & 0.3510  & 0.4811  &  0.3399 \\
 
     FIFO  & 0.9269  & 0.9774  & 0.9943  & 0.4971  & 0.6550  & 0.4763 & 0.3257  & 0.2807  & 0.1481  & 0.3609  & 0.4811  & 0.3399 \\
 
     \textit{GDumb}~\cite{prabhu2020gdumb}  &  0.9269   & 0.7076  & 0.9502  & 0.9683  & 0.8293  & 0.9767 & 0.6240  & 0.4704  & 0.6521  & 0.7734  & 0.6481  & 0.6628 \\
 
     TinyER~\cite{chaudhry2019tiny}  &  0.9852   & 0.9752  & 1.0064  & 0.9766  & 0.8584  & 0.9723 &   0.9324  & 0.7995  & 0.9315  & 0.8825  & 0.7074  & 0.8525 \\
 
 
     DER~\cite{buzzega2020dark}  &  0.5976  & 0.8625  & 0.9807  & 0.8727  & 0.8402  & 0.9734 & 0.7972  & 0.8397  & 0.9870  & 0.8293  & 0.8286  & 0.9630 \\
 
 
     DER++~\cite{buzzega2020dark}  & 0.9004   & 0.9020  & 0.9985  & 0.9398  & 0.8746  & 0.9786 & 0.8785  & 0.8547  & 0.9933   & 0.9125  & 0.8484  & 0.9696 \\
 
     CLS-ER~\cite{arani2022learning}  & 0.9573  & 0.1837  & 0.1107  & 0.5010  & 0.6664  & 0.3182 & 0.6854  & 0.1837  &  0.1107  & 0.5007  & 0.5858  & 0.2580 \\
 
 
 
 
     ExStream~\cite{hayes2019memory}  &  0.9114  & 0.8053  & 0.9286  & 0.9035  & 0.8375  & 0.8884 &  0.8713  & 0.7389  & 0.8530  & 0.8806  & 0.8339  & 0.9091 \\
       
 
     REMIND~\cite{hayes2019remind}  & 0.9666  & 0.9483  & 0.9988  &  0.9544  & 0.8197  & 0.9507 & 0.9102  & 0.7764  & 0.8993  & 0.8453  & 0.6784  & 0.8259 \\

     \textbf{Ours}  &  \textbf{1.0087} & \textbf{1.0045} & \textbf{1.0202} & \textbf{1.0069} & \textbf{0.8874} & \textbf{0.9918} &  \textbf{0.9613} & \textbf{0.8555} & \textbf{0.9945} & \textbf{0.9985} & \textbf{0.8840} & \textbf{0.9851} \\

     \midrule
 
     Offline  &  1.000  & 1.0000 & 1.0000 & 1.0000 & 1.0000 &  1.0000 &  1.000  & 1.0000 & 1.0000 & 1.0000 & 1.0000 &  1.0000 \\
 
     $\widehat{\text{Offline}}$  & 0.8046   & 0.8726 & 0.9038 & 0.8785 & 0.9266 & 0.9268 &  0.8046  & 0.8726 & 0.9038 & 0.8785 & 0.9266 & 0.9268 \\

    \bottomrule
   \end{tabular}
 
 }

 \vspace{-2em}
   
\end{table*}

\vspace{-0.4em}

The optimization above is a virtual/local gradient update, as it doesn't alter the model parameters $\theta$. However, $\theta^{v}$ is optimized focusing on the novel sample, which may not generalize well to the observed past samples due to changes in the previously optimal network parameters, leading to forgetting. To address this, we perform a global optimization with rehearsal. For this, we choose two more sample subsets from memory: $\mathcal{D}^{t}_{l}, \mathcal{D}_{\mathcal{M}}^{t} \sim \mathcal{M}$, each with size $C$. Let $H_{l}^{t} \leftarrow F_{\Phi}(\mathcal{D}_{l}^{t})$ represent the logits obtained over replay samples $\mathcal{D}^{t}_{l}$, with $F_{\Phi}(\cdot)$ denoting the semantic memory (see Sec.~\ref{sec:semantic_memory}). Then, we compute the loss over virtual parameters $\theta^{v}$ using both subsets from the replay buffer as the sum of cross-entropy and knowledge distillation loss~\cite{hinton2015distilling}, defined as:
\begin{equation}\label{eq:generalization_loss}
\small
    \mathcal{L}^{t} = \mathbb{E}_{(\boldsymbol{z}, y) \sim \mathcal{D}^{t}_{\mathcal{M}}} \left[ \mathcal{L}_{CE}(y, F_{\theta^{v}}(\boldsymbol{z})) \right] + \lambda \mathcal{L}_{MSE}(H^{t}_{l}, F_{\theta^{v}}(\mathcal{D}^{t}_{l}))
\end{equation}
Eq.~\ref{eq:generalization_loss} assesses the generalization loss over the rehearsal subsets using virtual parameters optimized for the new streaming example. If the model exhibits forgetting, then the virtual parameters will incur high loss due to poor generalization. Otherwise, the loss will be small.

Suppose $\nabla_{\theta^{v}}\mathcal{L}^{t}(F_{\theta^{v}})$ be the gradient of the loss in Eq.~\ref{eq:generalization_loss} w.r.t. the virtual model parameters $(\theta^{v})$. Then, we can compute the \emph{global parameters}, for the plastic network $(\theta)$, as follows:
\vspace{-0.1em}
\begin{equation}\label{eq:parameterupdate}
    \theta \leftarrow \theta - \beta \nabla_{\theta^{v}}\mathcal{L}^{t}(F_{\theta^{v}})
\end{equation}

Eq.~\ref{eq:parameterupdate} updates $\theta$ using the gradient of the virtual parameter, which might appear counter-intuitive. However, the alternating competitive training between Eq.~\ref{eq:virtualgrad} and Eq.~\ref{eq:parameterupdate} is crucial. Below, we briefly discuss this training behavior.

The updates in Eq.~\ref{eq:virtualgrad} and Eq.~\ref{eq:parameterupdate} occur alternately. Eq.~\ref{eq:virtualgrad} assesses how well $\theta$ generalizes to new data, while Eq.~\ref{eq:parameterupdate} focuses on $\theta^{v}$'s generalization to past data. For minimum loss (in Eq.~\ref{eq:lossnew}), $\theta$ must adapt to new samples, which is more challenging as compared to $\theta^{v}$, which is optimized for new samples. Conversely, $\theta^{v}$ may struggle to generalize to past samples due to its emphasis on new data. Hence, both $\theta$ and $\theta^{v}$ need to generalize effectively to new and replayed samples, minimizing losses in both Eq.~\ref{eq:lossnew} and Eq.~\ref{eq:generalization_loss}. In this alternating learning, convergence occurs when these losses approach zero, implying mutual generalization and reduced forgetting.

\vspace{-0.2em}
\subsection{Tiny Episodic Memory (TEM)}\label{sec:tiny_episodic_memory}
\vspace{-0.2em}

The model uses a fixed-sized tiny episodic memory (TEM) to act as short-term memory. In each incremental step, it $(i)$ replays subsets of $(C)$ samples uniformly selected from memory for continual learning, and $(ii)$ stores the new sample in memory. It employs Reservoir Sampling~\cite{vitter1985random} and Class Balancing Random Sampling to maintain a fixed-sized replay buffer. Reservoir Sampling selects a random buffer sample to replace with the new example, while Class Balancing Random Sampling picks a sample from the most populated class in the buffer to replace the new one.

\vspace{-0.2em}
\subsection{Semantic Memory (SEM)}\label{sec:semantic_memory}
\vspace{-0.2em}

Semantic memory (SEM) retains long-term knowledge and combats forgetting using self-distillation-loss~\cite{hinton2015distilling,allen2020towards,mobahi2020self,zhang2019your,zhang2021self} minimization (Eq.~\ref{eq:generalization_loss}), aligning the current model's decision boundary with past memories. SEM, based on a DNN and initialized with working model parameters, absorbs knowledge from the working network $\theta$ in incremental steps. Inspired by mean-teacher~\cite{tarvainen2017mean,cai2019exploring,deng2021unbiased,chen2020multi}, SEM is updated stochastically via exponential moving average (EMA) rather than at every iteration. Given a randomly sampled value $u$ from a uniform distribution ($u\sim \mathcal{U}(0,1)$) and an acceptance probability $r$, the update process for SEM denoted with $\Phi$ and the working model parameter $\theta$, is defined as follows
\vspace{-0.1em}
\begin{equation}\label{eq:phi_computation}
    \Phi = \begin{cases} 
                     \gamma \; \Phi + (1 - \gamma) \; \theta &, u < r \\
                     \Phi &, Otherwise \\
                   \end{cases}
\end{equation}
\vspace{-0.2em}
The acceptance probability is a hyper-parameter that regulates the frequency of SEM updates. A lower (higher) acceptance probability means less (more) frequent updates, retaining more (less) information from the remote model. This update resembles the mammalian brain, with information initially stored in short-term memory before transitioning to long-term memory. Algorithm~\ref{alg:proposed_model} illustrates the different stages of the proposed model.

\vspace{-0.5em}

\section{Experiments}\label{sec:experiments}
 \vspace{-0.3em}
 
\subsection{Datasets, Data Orderings and Metrics}\label{sec:datasets_and_metrics}
 
\textbf{Datasets.} We evaluate our method ({\pname}) through extensive experiments on three temporally coherent datasets: iCub1.0~\cite{fanello2013icub}, iCub28~\cite{pasquale2015teaching}, and CoRe50~\cite{lomonaco2017core50}. iCub1.0 involves object recognition from video frame sequences, with each frame containing a single object instance, while iCub28 is similar but spans across four days. CoRe50, like iCub1.0/28, includes temporally ordered images divided into 11 sessions with varying backgrounds and lighting.

\textbf{Data Orderings.} To test {\pname}'s robustness in a challenging SLL setup, we assess its streaming learning capability using four data-ordering schemes, as in~\cite{hayes2019remind,hayes2019memory,banerjee2023streaming}. These schemes include (i) streaming i.i.d., (ii) streaming class i.i.d., (iii) streaming instance, and (iv) streaming class instance ordering.

\vspace{-0.3em}

\begin{algorithm}[h]
  

  \caption{\textbf{{\pname}}}\label{alg:proposed_model}
  \begin{algorithmic}[1]
  \Require Initialize: $\Phi = \theta$ 
  \Require Hyperparameters: $\lambda, r, \alpha, \beta, \gamma$, Memory: $\mathcal{M}$
  
  \For{$t \in 1, \dots, T, \dots$} 
    
    \State $\{(\boldsymbol{z}_{t}, y_{t})\} = \{(G_{\xi}(\boldsymbol{x}_{t}), y_{t})\} \sim \mathcal{D}_{t}$ \Comment{$|\mathcal{D}_{t}| = 1$}

    \State Select samples from $\mathcal{M}$: $\mathcal{D}_{v}^{t} \sim \mathcal{M}$  \Comment{$|\mathcal{D}_{v}^{t}| = C$}

    \State $\mathcal{D}_{v}^{t} \leftarrow \mathcal{D}_{t} \cup \mathcal{D}_{v}^{t}$

    \State Compute $\mathcal{L}^{t}_{v}$ using Eq.~\ref{eq:lossnew} and evaluate $\nabla_{\theta}\mathcal{L}^{t}_{v}(F_{\theta})$

    \State $\theta^{v} = \theta - \alpha \nabla_{\theta}\mathcal{L}^{t}_{v}(F_{\theta})$ \Comment{Virtual gradient update}

    \State Select samples from $\mathcal{M}$: $\mathcal{D}^{t}_{l}, \mathcal{D}_{\mathcal{M}}^{t} \sim \mathcal{M}$  \Comment{$|\mathcal{D}_{l}^{t}| = |\mathcal{D}_{\mathcal{M}}^{t}| = C$}
    
    \State $H_{l}^{t} \leftarrow F_{\Phi}(\mathcal{D}_{l}^{t})$ 

    \State Compute $\mathcal{L}^{t}$ using Eq.~\ref{eq:generalization_loss} and evaluate $\nabla_{\theta^{v}}\mathcal{L}^{t}(F_{\theta^{v}})$ 

    \State $\theta = \theta - \beta \nabla_{\theta^{v}}\mathcal{L}^{t}(F_{\theta^{v}})$ \Comment{Global update}

    \State sample $u \sim \mathcal{U}(0, 1)$
    
    \If{$u < r$}

      \State $\Phi \leftarrow \gamma \; \Phi + (1 - \gamma) \; \theta$ 

    \EndIf

    \State $\textit{UpdateMemory}(\mathcal{M}, \mathcal{D}_{t}, t)$ \Comment{Add sample to $\mathcal{M}$}

  \EndFor

  \State \Return $\theta, \Phi$

  \end{algorithmic}


\end{algorithm}

\begin{table}[t]

   \vspace{-0.7em}
   \scriptsize
   \centering
   
   \caption{Buffer capacity used for various datasets.}
   \label{tab:Table_memory_buffer_capacity}
   \scriptsize
   \vspace{-0.5em}
   \resizebox{0.45\textwidth}{!}{
   \begin{tabular}{ c | c | c | c | c } 
      \toprule
   
      \textbf{Dataset} & iCub1.0 & iCub28 & CoRe50 & ImageNet100 \\
   
      \midrule
   
      \textbf{Buffer Capacity} & 230 & 230 & 1000 & 1000 \\
   
      \midrule
   
      \textbf{Training-Set Size} & 6002 & 20363 & 119894 & 127778 \\
      
   \bottomrule
   \end{tabular}
   
   }
   
   \vspace{-2em}
   
\end{table}

\textbf{Metrics.} To assess the learner's performance in a CISLL setup, we employ the $\boldsymbol{\Omega}_{\text{all}}$ metric, following the approach in~\cite{kemker2017fearnet,hayes2019memory,hayes2019remind,banerjee2023streaming}. This metric quantifies CL performance normalized against an \emph{Offline} baseline: 
\vspace{-0.3em} 
\begin{equation}\label{eq:metric}
\boldsymbol{\Omega}_{\text{all}} = \frac{1}{T} \sum^{T}_{t = 1} \frac{\boldsymbol{\alpha}_{t}}{\boldsymbol{\alpha}_{\text{Offline}, t}}
\end{equation}
\vspace{-0.3em}
\noindent where $(i)$ $T$ is the total number of testing events, $(ii)$ $\boldsymbol{\alpha}_{t}$ is the streaming learner's unnormalized performance at time $t$, and $(iii)$ $\boldsymbol{\alpha}_{\text{Offline}, t}$ is the unnormalized performance of the Offline baseline at time $t$.

\begin{figure*}[ht]

  \centering
  \includegraphics[width=17cm, height=7cm]{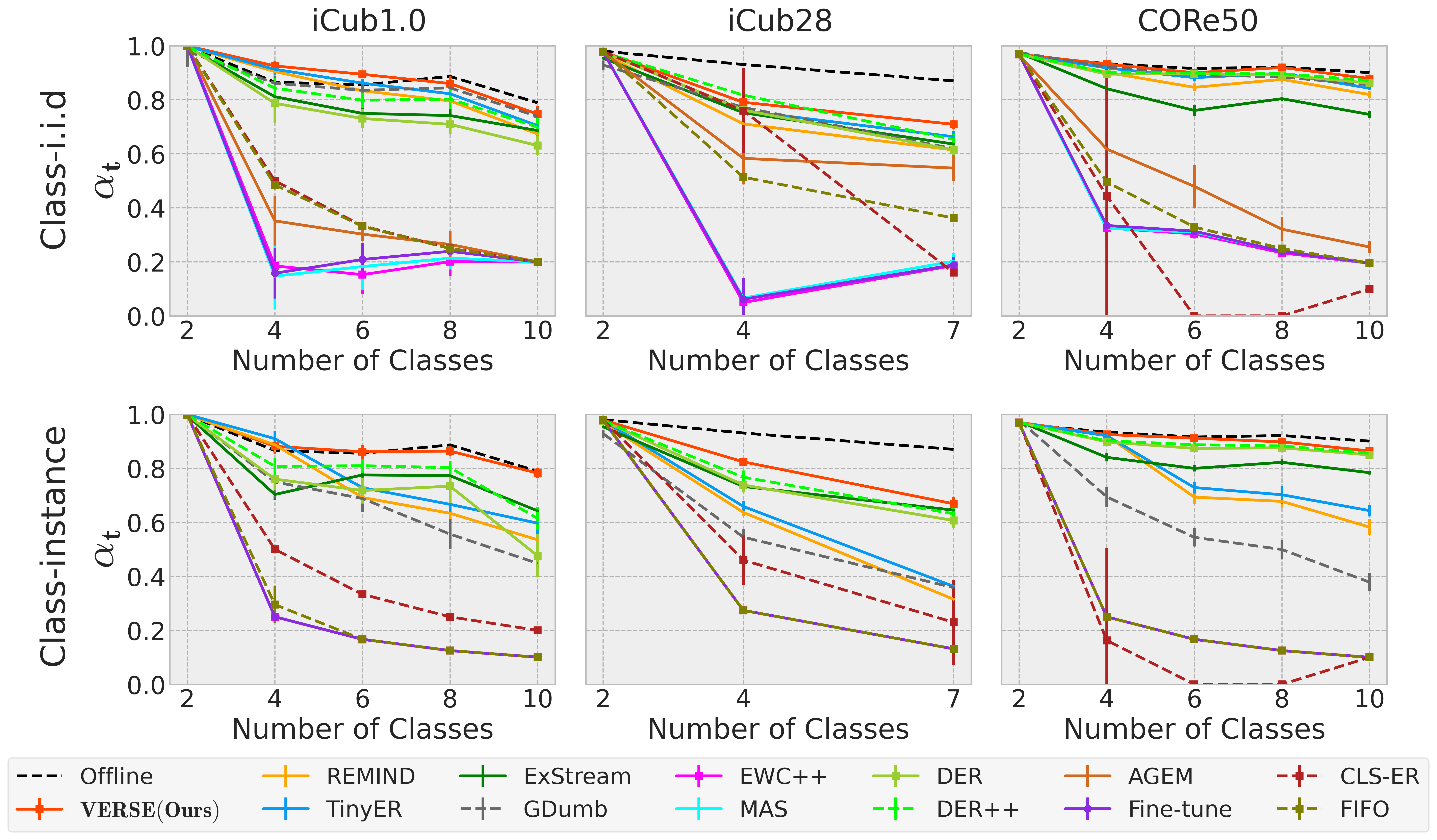}
  
 \vspace{-0.5em}

  \caption{Plots of $\boldsymbol{\alpha}_{t}$ as a function of streaming learning model and data-orderings. {\pname} outperforms other SLL models in both streaming class-iid (top-row) and streaming class-instance (bottom-row) orderings across datasets.}
  \label{fig:icubworld_1_0_alpha_plots}
 \vspace{-1.8em}
\end{figure*}

\subsection{Baselines and Compared Methods}\label{sec:baselines_and_compared_methods}

{\pname} adheres to the challenging CISLL approach. We compare it with ExStream~\cite{hayes2019memory} and REMIND~\cite{hayes2019remind}. We also evaluate various IBL and OCL approaches (EWC++, MAS, AGEM, GDumb, TinyER, DER/DER++, CLS-ER, FIFO), as well as two additional baselines: $(i)$ offline training with full dataset access (Offline/Upper Bound), and $(ii)$ fine-tuning with one example at a time and no CL strategy (Fine-tuning/Lower Bound). All comparisons are performed under the SLL setup, except for \textit{GDumb}, which fine-tunes with replay buffer samples, giving it an unfair advantage.

\begin{figure}[t]
  \centering
  
  \vspace{0.4em}

  \includegraphics[width=8.5cm, height=4cm]{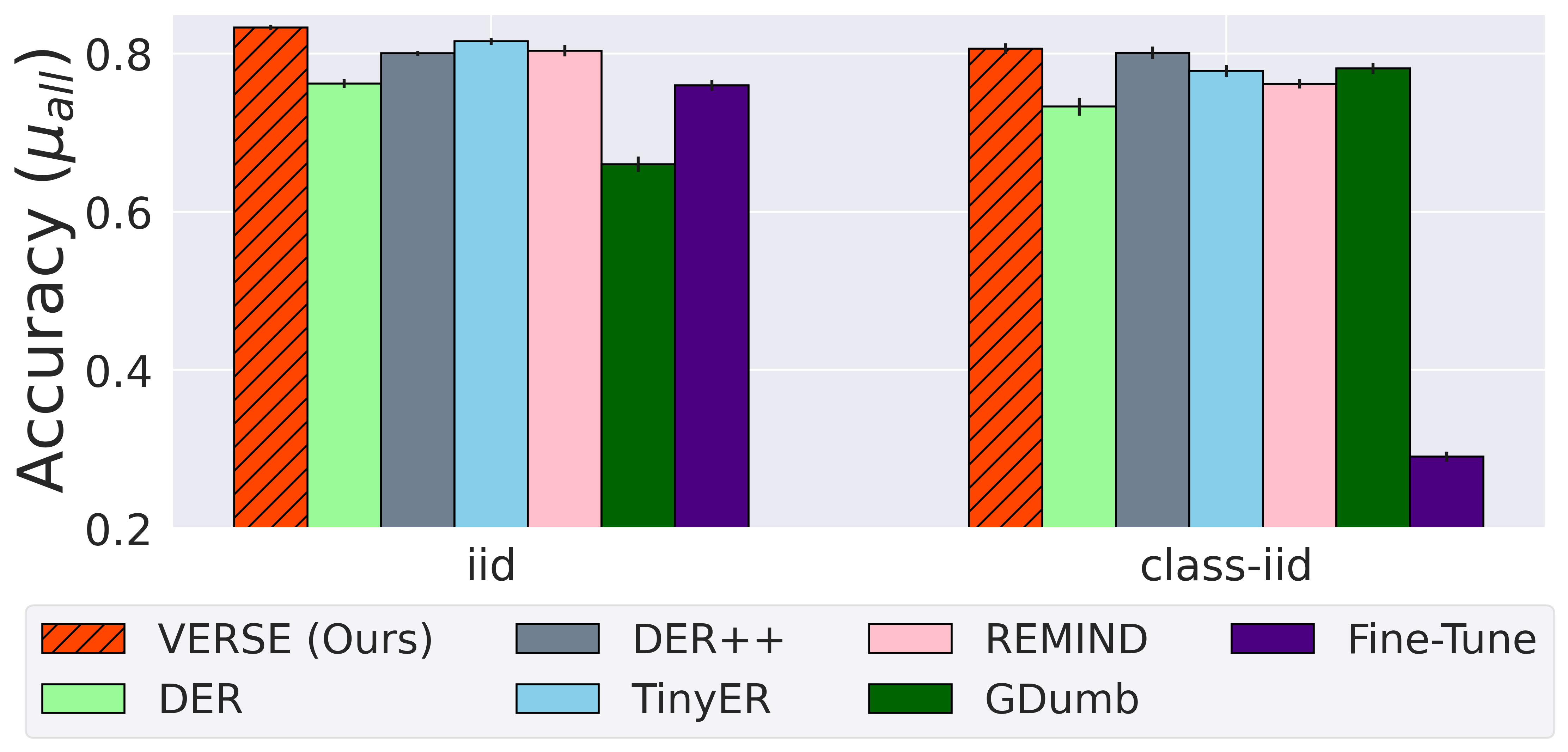}
  
 \vspace{-0.5em}

  \caption{Performance ($\boldsymbol{\mu}_{\text{all}}$) comparison between {\pname} (Ours) and the other baselines on ImageNet100.}
  \label{fig:imagenet100}
 \vspace{-1.8em}
\end{figure}

\subsection{Implementation Details}\label{implementation_details}

In all experiments, baselines are trained with one sample at a time using the same network architecture. We employ ResNet-18~\cite{he2016deep} pretrained on ImageNet-1K~\cite{deng2009imagenet,russakovsky2015imagenet} available in PyTorch~\cite{paszke2019pytorch} TorchVision package, using its first 15 convolutional (conv) layers and 3 downsampling layers as the feature extractor $(G)$. The remaining 2 conv layers and 1 fully connected (FC) layer constitute the plastic network $(F)$. For ExStream~\cite{hayes2019memory}, all 17 conv and 3 downsampling layers are utilized for feature extraction $(G)$, and the final FC layer serves as the plastic network $(F)$. Feature embeddings are stored in memory for all baselines, including {\pname}. Replay buffer capacity is specified in Table~\ref{tab:Table_memory_buffer_capacity}. We employ reservoir sampling for class-instance and instance ordering and class-balancing random sampling for class-iid and iid ordering. Experience-replay and self-distillation consistently use $C = 16$ samples across all baselines. Hyperparameters are set as follows: $\alpha = 0.005, \beta = 0.01, \lambda = 0.3$, and $\gamma = 0.9$. $r$ values are set as: $(i)$ $r = 0.4$ for iCub1.0, $(ii)$ $r = 0.1$ for iCub28, and $(iii)$ $r = 0.05$ for CORe50 dataset. Each experiment is repeated 10 times with different data permutations, and the average accuracy is reported.

\begin{figure}[t]
  \centering
  
  \vspace{0.4em}

  \includegraphics[width=8.6cm, height=3.6cm]{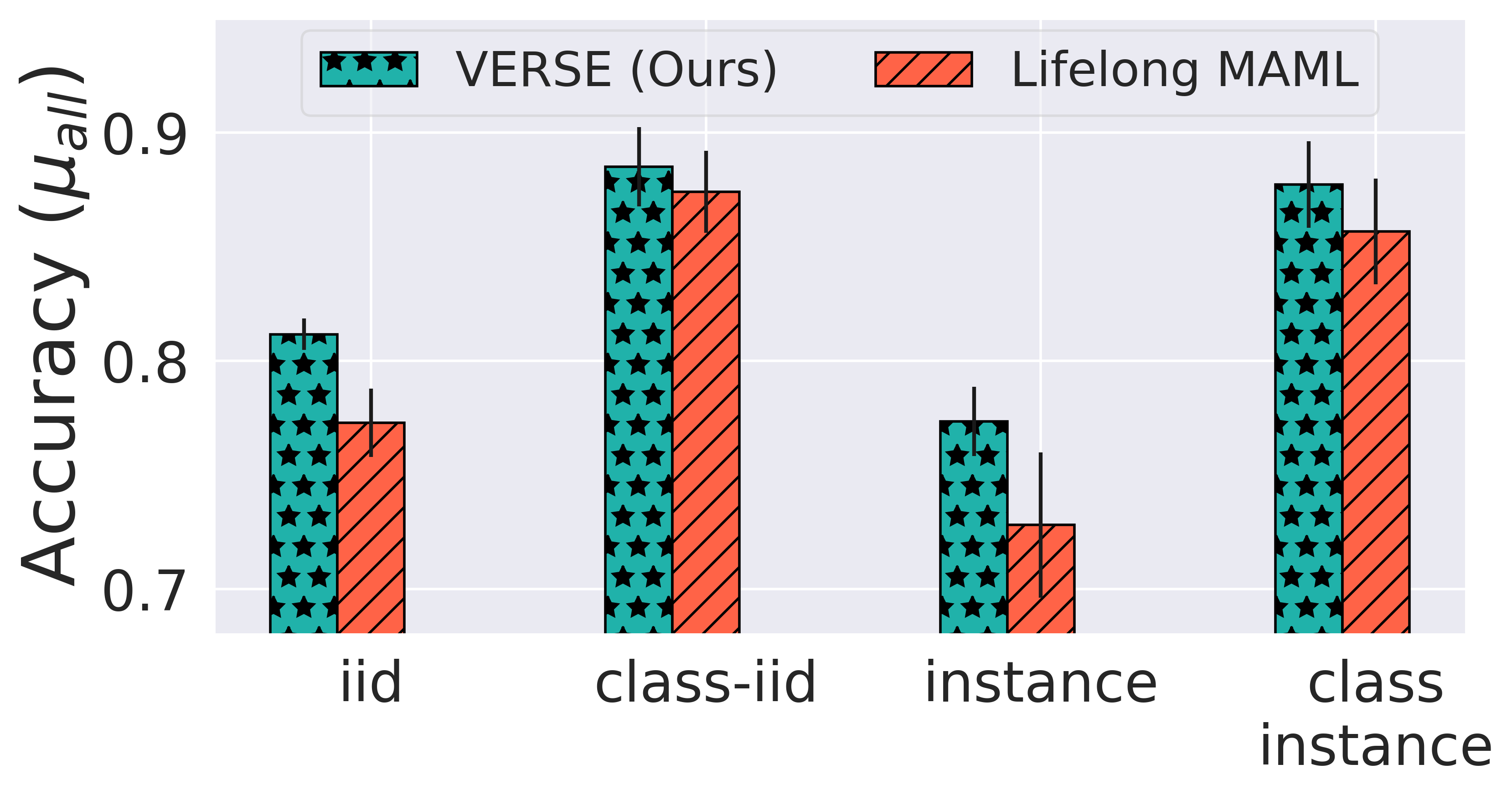}
  
 \vspace{-0.5em}

  \caption{Performance ($\boldsymbol{\mu}_{\text{all}}$) comparison between {\pname} and Lifelong MAML~\cite{gupta2020look} on iCub1.0.}
  \label{fig:comparison_with_lifelong_maml}
 \vspace{-1.8em}
\end{figure}

\begin{figure*}[ht]
    \centering
    
    \includegraphics[width=17cm, height=3.2cm]{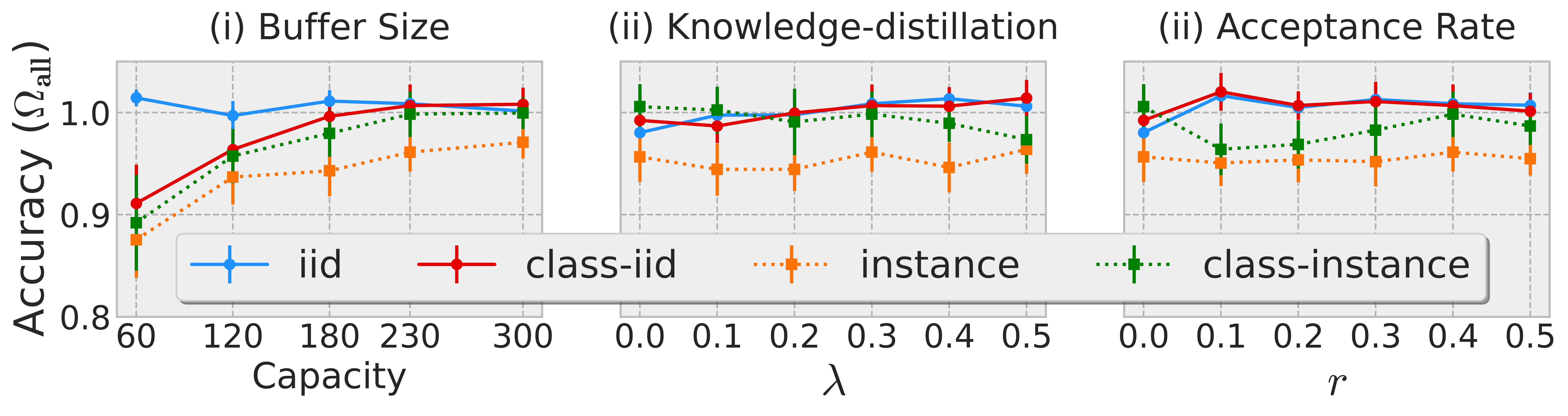}  
   \vspace{-0.5em}
  
    \caption{Plots of $\boldsymbol{\Omega}_{\text{all}}$ as as function of $(i)$ replay buffer capacity $(|\mathcal{M}|)$, $(ii)$ knowledge-distillation hyper-parameter $(\lambda)$, and $(iii)$ acceptance-rate hyper-parameter $(r)$ on iCub1.0.} 
    \label{fig:icubworld_1_0_ablation_studies}
   \vspace{-1.8em}
\end{figure*}

\subsection{Results}\label{sec:results}

Table~\ref{Table_omega_all_results} presents {\pname}'s performance in various experimental setups with different data orderings and datasets. We conducted 10 repetitions of each experiment, reporting average accuracy. Notably, {\pname} consistently surpasses the baselines by a significant margin. It demonstrates robustness to different data-ordering schemes, which are known to induce catastrophic forgetting. In contrast, IBL methods like EWC++~\cite{kirkpatrick2017overcoming,chaudhry2018riemannian} and MAS~\cite{aljundi2018memory} experience severe forgetting. Even \textit{GDumb}~\cite{prabhu2020gdumb}, which fine-tunes network parameters with buffer samples before each inference, fails to outperform {\pname}.

iCub1.0/28 and CoRe50 are temporally coherent datasets, offering a more realistic and challenging evaluation scenario. Class-instance and instance ordering requires the agent to learn from temporally ordered video sequences one at a time. Table~\ref{Table_omega_all_results} shows that {\pname} achieves notable improvements: $(i)$ up to $8.6\%$ and $2.89\%$ on iCub1.0 for class-instance and instance ordering, $(ii)$ $3.56\%$ on iCub28 for class-instance ordering, and $(iii)$ $1.55\%$ on CoRe50 for class-instance ordering. Fig.~\ref{fig:icubworld_1_0_alpha_plots} plots accuracy ($\alpha_{t}$) of {\pname} (Ours) and other baselines for class-iid and class-instance ordering. Notably, {\pname} better retain knowledge of old classes compared to other baselines, particularly excelling in class-instance ordering.

We also assess {\pname} and other baselines on ImageNet100, a subset of ImageNet-1K (ILSVRC-2012)~\cite{deng2009imagenet,russakovsky2015imagenet}, comprising randomly selected 100 classes, each with 700-1300 training samples and 50 validation samples. As ImageNet-1K lacks labels for test samples, we used the validation set for testing, following~\cite{hayes2019remind}. Fig.~\ref{fig:imagenet100} illustrates the performance  ($\boldsymbol{\mu}_{\text{all}}$) of various baselines, including {\pname}, showing {\pname} consistently outperforming all other CL methods. $\boldsymbol{\mu}_{\text{all}}$ represents the mean-absolute accuracy with $  \boldsymbol{\mu}_{\text{all}} = \frac{1}{T} \sum^{T}_{t = 1} \boldsymbol{\alpha}_{t}$,
where $(i)$ $T$ is the total number of testing events and $(ii)$ $\boldsymbol{\alpha}_{t}$ is the accuracy of the streaming learner at time $t$.

Finally, we compare {\pname} and Lifelong MAML~\cite{gupta2020look}, a continual learning variant of MAML~\cite{finn2017model}, on the iCub1.0 dataset. Fig.~\ref{fig:comparison_with_lifelong_maml} illustrates the comparison using $\boldsymbol{\mu}_{\text{all}}$ metric with {\pname} consistently outperforming LifeLong MAML across all data-orderings.

\begin{figure}[t]
  \centering
  
  \includegraphics[width=7.2cm, height=3.2cm]{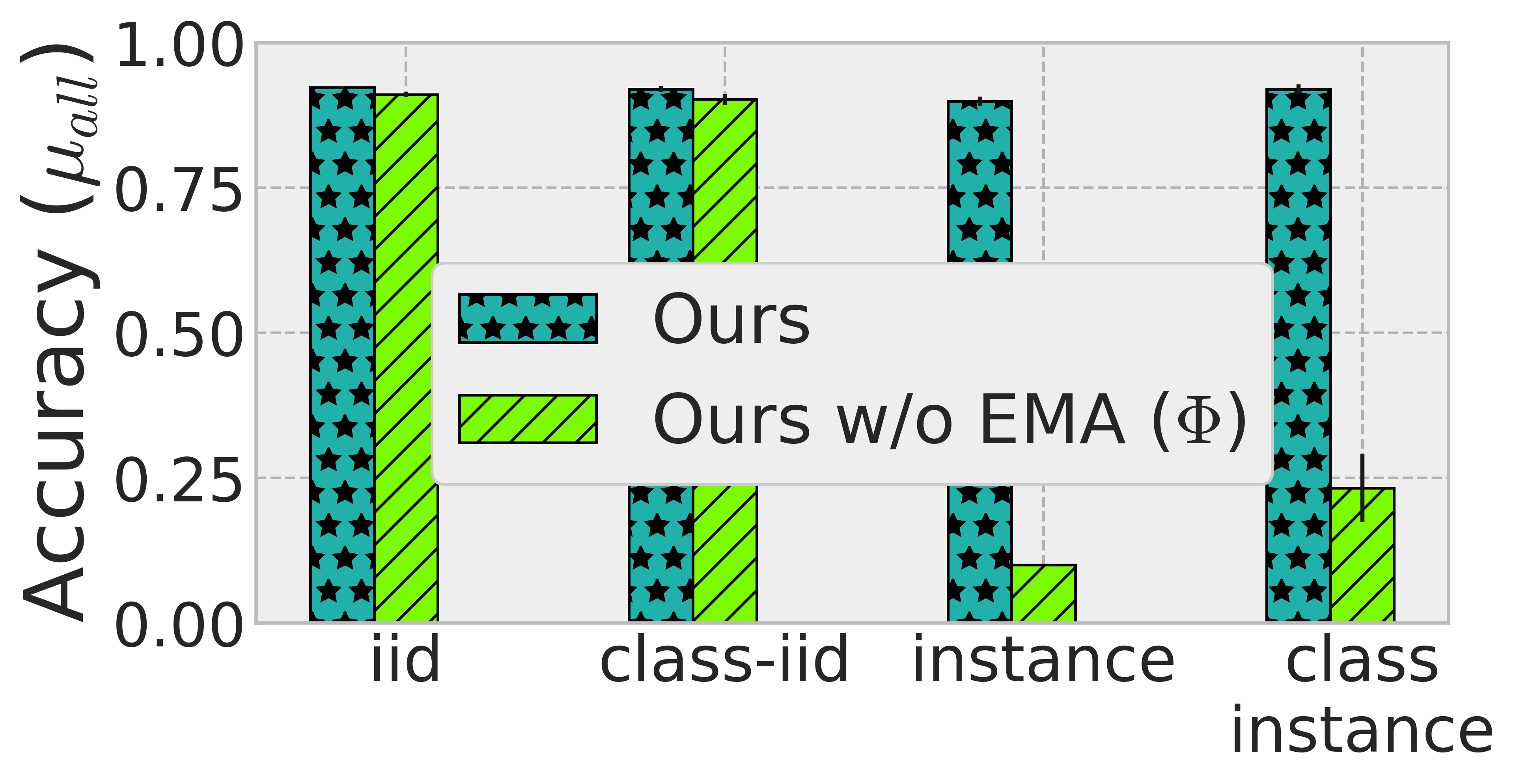}
  
 \vspace{-0.5em}

  \caption{Plots of $\boldsymbol{\mu}_{\text{all}}$ as a function of EMA $(\Phi)$ used to compute the semantic memory (Eq.~\ref{eq:phi_computation}) on CORe50.}
  \label{fig:ablations_on_core50}
 \vspace{-1.8em}
\end{figure}

\vspace{-0.3em}
\section{Ablations}\label{sec:ablations}

We perform extensive ablations to validate the importance of the various components of {\pname}.

\textbf{Choice Of Buffer Capacity.} Fig.~\ref{fig:icubworld_1_0_ablation_studies} (left) shows the impact of different buffer capacities on iCub1.0. Increased buffer capacity leads to improved model performance.

\textbf{Choice Of Hyper-parameter $(\lambda)$.} Fig.~\ref{fig:icubworld_1_0_ablation_studies} (middle) shows the impact of changing the self-distillation hyper-parameter $(\lambda)$ on iCub1.0. The best performance is consistently achieved across all data orderings with $\lambda = 0.3$.

\textbf{Significance Of Self-Distillation Loss.} Fig.~\ref{fig:icubworld_1_0_ablation_studies} (middle) depicts the model's performance with $\lambda = 0.0$, indicating no self-distillation. While the best performance is achieved with $\lambda = 0.3$, self-distillation alone does not significantly improve performance.

\textbf{Significance Of Acceptance-Rate $(r)$.} Fig.~\ref{fig:icubworld_1_0_ablation_studies} (right) illustrates the impact of changing the acceptance-rate $(r)$ on iCub1.0. The best performance is achieved with $r = 0.4$. However, increasing $r$ to $0.50$ leads to performance degradation. For instance ordering, the model tends to perform best with $r = 0.0$.

\textbf{Significance of Exponential Moving Average (EMA).} Fig.~\ref{fig:ablations_on_core50} highlights the importance of SEM $(\Phi)$ in the model's performance. Without using SEM and relying solely on the working model $(\theta)$ for computing logits in self-distillation loss (Eq.~\ref{eq:generalization_loss}), the model's performance degrades. Additionally, for temporally coherent orderings (instance and class instance orderings), not using EMA to update SEM severely degrades {\pname}'s performance.

\begin{figure}[t]
  \centering
  
  \includegraphics[width=7.5cm, height=3.2cm]{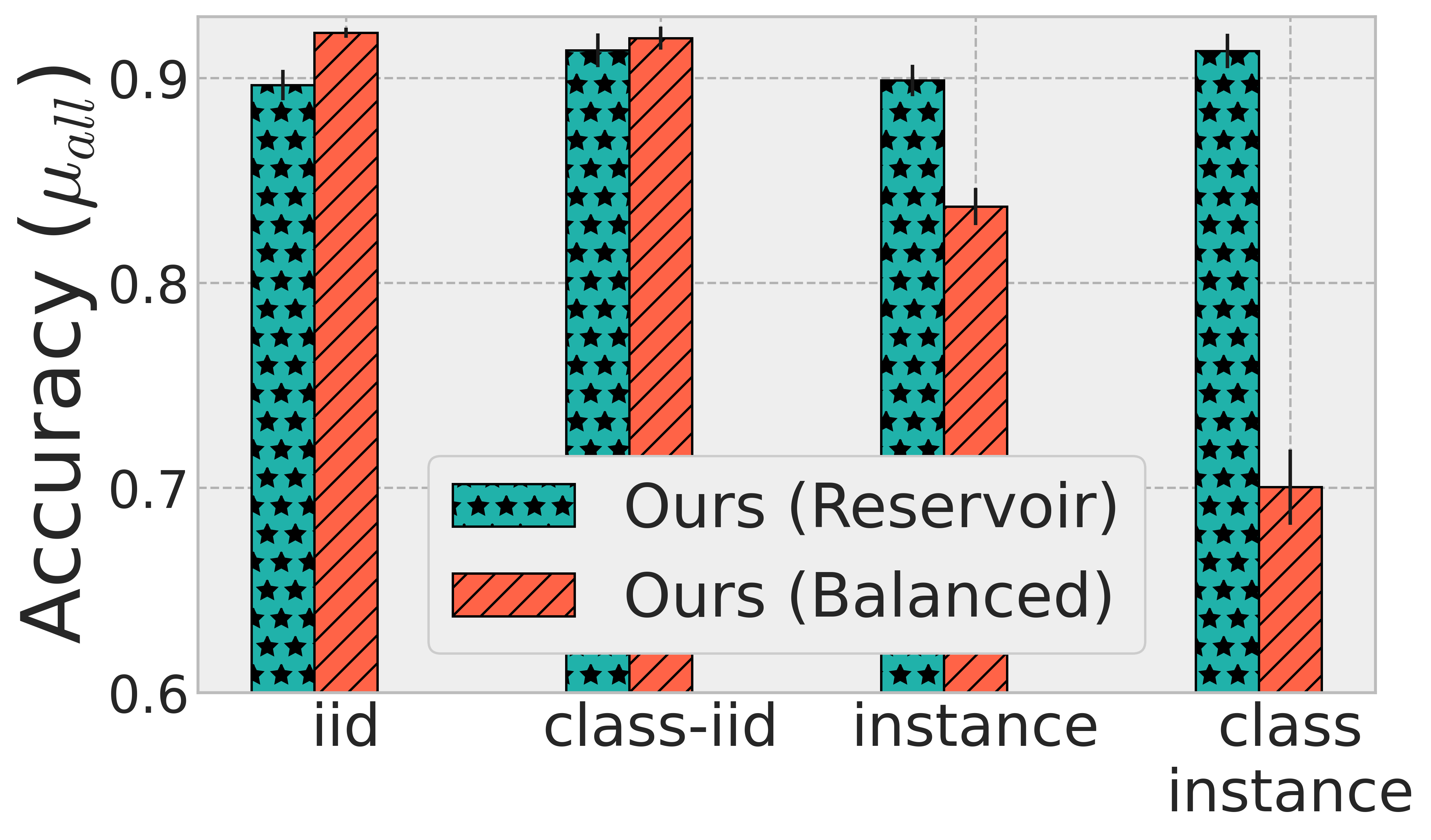}
  
 \vspace{-0.5em}

  \caption{Plots of $\boldsymbol{\mu}_{\text{all}}$ as a function of buffer replacement policies on iCub1.0.}
  \label{fig:ablations_buffer_replacement}
 \vspace{-1.8em}
\end{figure}

\textbf{Significance of Buffer Replacement Policies.} Fig.~\ref{fig:ablations_buffer_replacement} shows the model's performance with different buffer replacement policies used for TEM. For temporally ordered data (instance and class instance ordering), reservoir sampling yields the best performance. However, for i.i.d and class i.i.d ordering, class balancing random sampling or balanced sampling achieves the best results.

\section{Conclusion}\label{conclusions}

We address the challenging problem of streaming lifelong learning, where the learner is given only one sample at a time during training, the learned model is required to have anytime inference capability. Our replay-based virtual-gradient-regularization with global and virtual/local parameters generalization to both previous and novel task samples. Tiny episodic memory for rehearsal and semantic memory help align the decision boundary with past memories through self-distillation-loss. Extensive experiments and ablations on various datasets and data orderings demonstrate our approach's efficacy.

\clearpage

\medskip
{\small
\bibliographystyle{IEEEtran}
\bibliography{ref.bib}
}

\end{document}